\documentclass[letterpaper, 10 pt, conference]{ieeeconf}
\IEEEoverridecommandlockouts  
\usepackage{cite}
\usepackage{amsmath,amssymb,amsfonts}
\usepackage{algorithmic}
\usepackage{graphicx}
\usepackage{textcomp}
\usepackage{xcolor}
\usepackage{mathrsfs}
\usepackage{amssymb}
\usepackage[flushleft]{threeparttable}
\usepackage{booktabs}
\usepackage[font=small,skip=5pt]{caption}
\usepackage[font=small,position=b,skip=5pt]{subcaption}
\usepackage{multirow}
\usepackage[]{algorithm2e}

\usepackage{xparse} 
\usepackage{wrapfig}
\usepackage{hyperref}
\usepackage[capitalise]{cleveref}
\usepackage{dsfont}
\usepackage{comment}
\usepackage{color}
\usepackage{float}
\usepackage[bottom]{footmisc}

\NewDocumentCommand\bbm{}{ \begin{bmatrix} }
\NewDocumentCommand\ebm{}{ \end{bmatrix} }

\NewDocumentCommand\Norm{m}{\left\Vert#1\right\Vert }

\NewDocumentCommand\Real{}{ \mathbb{R} }

\NewDocumentCommand\LieGroupSE{m}{ \mathrm{SE}(#1) }

\NewDocumentCommand\AbsoluteValue{m}{ \left\vert#1\right\vert }

\NewDocumentCommand\Normal{}{\tilde{\mathbf{n}}}

\title{\textbf{Self-Supervised Scale Recovery for\\ Monocular Depth and Egomotion Estimation}}
\author{Brandon Wagstaff and Jonathan Kelly$^\dagger$
	\thanks{The authors are with the Space \& Terrestrial Autonomous Robotic Systems (STARS) Laboratory at the University of Toronto Institute for Aerospace Studies (UTIAS), Toronto, Ontario, Canada, M3H~5T6. Email: \texttt{<first name>.<last name>@robotics.utias.utoronto.ca}}
	\thanks{$^\dagger$Jonathan Kelly is a Vector Institute Faculty Affiliate. This research was supported in part by the Canada Research Chairs program.}}

\begin{document}
\makeatletter

\maketitle
\begin{abstract}
	The self-supervised loss formulation for jointly training depth and egomotion neural networks with monocular images is well studied and has demonstrated state-of-the-art accuracy. One of the main limitations of this approach, however, is that the depth and egomotion estimates are only determined up to an unknown scale. In this paper, we present a novel \textit{scale recovery loss} that enforces consistency between a known camera height and the estimated camera height, generating metric (scaled) depth and egomotion predictions. %
	We show that our proposed method is competitive with other scale recovery techniques that require more information. Further, we demonstrate that our method facilitates network retraining within new environments, whereas other scale-resolving approaches are incapable of doing so. Notably, our egomotion network is able to produce more accurate estimates than a similar method which recovers scale at test time only.
\end{abstract}

\section{Introduction}
Visual odometry (VO), or visual egomotion estimation, is a well-studied topic with a rich history \cite{Aqel:2016}. One of the known difficulties with monocular VO specifically, however, is that the true scale of the scene (relative to a known reference) cannot be resolved. Consequently, both the estimated scene depths and the estimated distance travelled between adjacent image frames are only determined up to an unknown scale factor. Also, this scale factor is prone to drift over time, and so a constant scale factor correction is not usually appropriate. Although simultaneous localization and mapping (SLAM) algorithms can mitigate scale drift through loop closure detection, extreme scale drift may cause loop closure to fail and may lead to irreversible errors during the map-building process \cite{frost2016object}.

While several existing techniques aim to resolve the scale factor ambiguity by incorporating scene information that is a priori known (e.g., known object sizes, or known camera height above the ground plane) into classical VO algorithms, this type of hand-engineering is challenging to tune for performance \cite{frost2018recovering}. An alternative solution is to resolve the metric scale factor through data-driven learning. Neural networks are able to map directly from raw image pairs to inter-frame egomotion predictions and, when ground truth labels (e.g., depths or poses) are available, can produce metrically-scaled outputs \cite{yin2017scale,kreuzig2019distancenet,Wang:2017}. However, since acquiring ground truth labels is often onerous and expensive, the amount of training data is usually limited. Furthermore, because ground truth training labels are not available at test time, newly-collected data cannot be used to update the network weights through online retraining. Learning-based systems are known to be unreliable outside of their training distribution \cite{krueger2020out} and hence the ability to update the network through retraining is important to enable long-term autonomy. For these reasons, self-supervised methods \cite{zhou:2017} have been proposed that jointly train depth and egomotion networks without requiring ground truth.

In the self-supervised loss formulation, a photometric reconstruction loss is employed during training. Although the self-supervised paradigm has evolved significantly recently, the network outputs remain \textit{unscaled}. This is because there is no metric information (e.g., from depth or pose labels) available during the training process.
Herein, we propose to use a \textit{scale recovery loss} that resolves metric scale by ensuring that the estimated camera height (over the ground plane) is the same as the known camera height. To the best of the authors' knowledge, our self-supervised learning-based system is the first to be able to produce scaled (metric) depth and egomotion estimates while only requiring monocular (as opposed to stereo) images during training. To enable the use of our novel scale recovery loss, we extract the ground plane from each training image and determine its normal and offset (i.e., camera height) through a least squares technique. The scale recovery loss then forces the estimated camera height to be consistent with the known camera height. By doing so, we inject metric information during the training process, which in turn causes the depth and egomotion networks to produce metrically scaled predictions that remain properly scaled at test time. Importantly, no ground plane segmentation is required at test time, unlike existing scale recovery methods \cite{xue2020toward}. In summary, our main contributions are:
\begin{figure*}[]
	\centering
	\begin{subfigure}[]{0.59\textwidth}
		\centering
		\includegraphics[width=\textwidth]{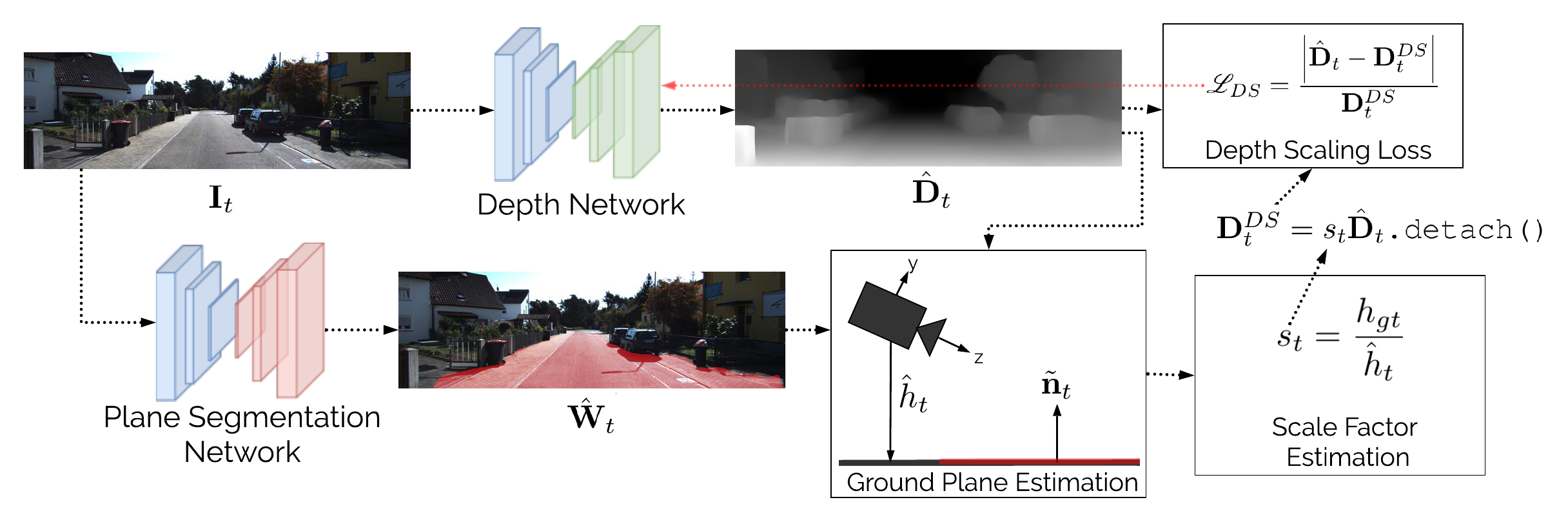}
		\caption{Depth scaling loss.}
		\label{fig:depth_scaling}
	\end{subfigure} 
	\begin{subfigure}[]{0.39\textwidth}
		\centering
		\includegraphics[width=\textwidth]{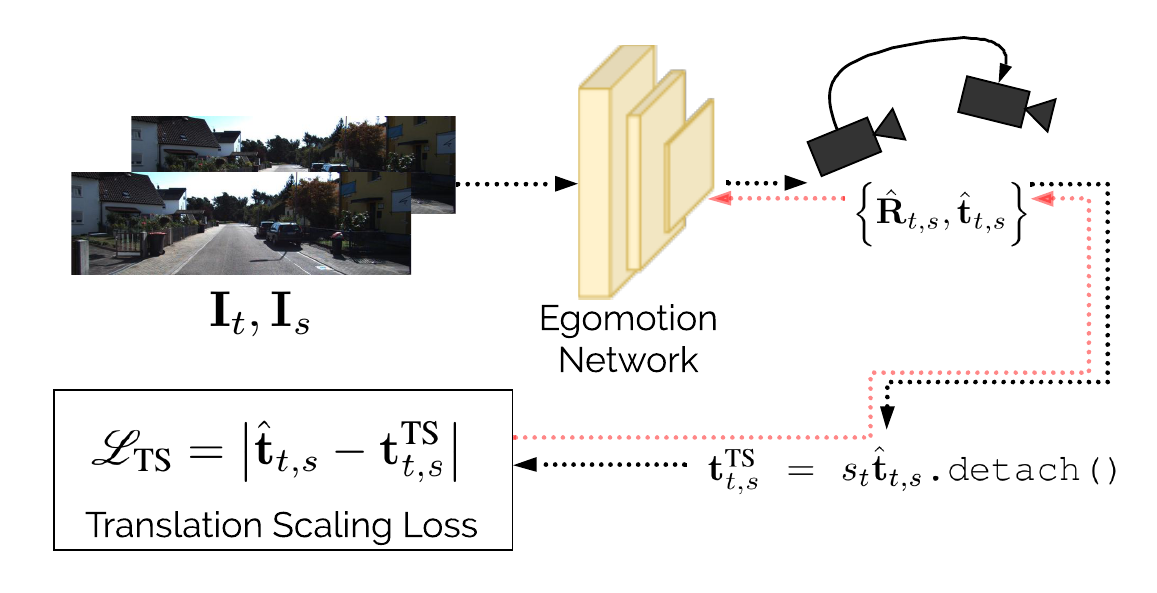}
		\caption{Translation scaling loss.}
		\label{fig:trans_scaling}
	\end{subfigure} 
	\caption{Illustration of our novel scale recovery loss formulation. For each training image, the scale factor is computed by comparing the estimated camera height to the known camera height; by enforcing the scale factor to converge to unity during training (through our proposed scale recovery loss), the network predictions become metrically scaled.}
	\label{fig:overview}
	\vspace{-0.2cm}
\end{figure*}
\begin{enumerate}
	\item a framework for training a self-supervised ground plane segmentation network,
	\item a novel loss function that enforces metrically-scaled depth and egomotion estimates without requiring ground truth labels or stereo images during training,
	\item comprehensive experiments showing that our loss functions can be used to train depth and egomotion networks to regress metrically-scaled predictions and furthermore can facilitate network retraining with a small amount of data collected online, and
	\item an open source implementation of our algorithm.\footnote{See \url{https://github.com/utiasSTARS/learned_scale_recovery} for supplementary material and our open source code archive.}
\end{enumerate}

\section{Related Work}

For monocular VO, accurate scale recovery, that is, the process of making relative depth and egomotion predictions consistent with metric, ground truth measurements, remains an active area of research. In this section, we describe existing methods in the literature for both classical and learning-based egomotion estimation.

\subsection{Scale Recovery in Classical Systems}

Lack of metric scale and scale drift are known problems for classical monocular VO estimators. Indeed, popular systems such as \texttt{ORB-SLAM2} \cite{mur2017orb} are prone to scale drift. The VO subsystem of \texttt{ORB-SLAM2} relies on bundle adjustment and loop closure to enforce an approximately constant scale factor over the complete trajectory, but is not able to resolve metric scale. 

Several methods \cite{song2015high, fanani2017multimodal, zhou2019ground, wang2018monocular, kitt2011monocular} attempt to determine the metric scale at test time by detecting the ground plane and comparing the estimated camera height (relative to the ground plane) with the known camera height. We draw inspiration from these methods but note that they have some key limitations. Many of the algorithms \cite{song2015high, zhou2019ground, kitt2011monocular} assume that the ground plane appears within a pre-defined image region, which is problematic when the ground plane is not visible (e.g., when the ground plane is occluded by a vehicle on the road). An alternative \cite{lee2015online} is to classify ground plane pixels using colour information, but because the hue and intensity of the ground plane pixels may change significantly depending on scene illumination and camera settings, this form of road plane detection is unreliable.
Wang et al.\ \cite{wang2018monocular} address this shortcoming by detecting the ground plane by fitting a model to 3D feature points. Although this technique is more robust to ground plane pixel hue and intensity changes, the main drawback is that the ground plane (being smooth and textureless) often lacks readily-identifiable features. To mitigate these difficulties, our ground plane segmentation network is trained using a geometry-based loss, which is independent of pixel intensity and illumination. Since we use a dense set of pixels to determine the ground plane, we expect to outperform feature-based plane detection in regions that lack identifiable features. Additionally, while existing methods \cite{song2015high, fanani2017multimodal, zhou2019ground, wang2018monocular, kitt2011monocular,lee2015online} require the presence of a visible ground plane in \textit{every} image at test time, our method only requires the presence of the ground plane in the training images.

\subsection{Scale Recovery in Learning-Based Systems}

The most straightforward means of enforcing metrically-scaled depth and egomotion predictions is through supervised learning (see \cite{Wang:2017,wang2019improving,costante2018ls}). However, collecting ground truth data can be time consuming, expensive, and it may not always be reliable (e.g., due to GNSS errors within urban canyons). Additionally, relying on ground truth limits the ability of learning-based systems (in our case, deep networks) to be retrained online in areas where ground truth is not available. Online retraining is important when deploying robots that must operate in environments which differ from the original training environment, motivating the use of self-supervised training methods.

Self-supervised learning of depth and egomotion, initially proposed in \cite{zhou:2017}, has become a prevalent approach, and recent work has demonstrated \cite{godard2019digging,kumar2019fisheyedistancenet,yang2020d3vo} near state-of-the-art accuracy for dense depth prediction from monocular images (while, in general, learning-based egomotion networks have not surpassed the accuracy of classical techniques). These systems are trained with a self-supervised photometric reconstruction loss along with a variety of secondary losses. To compute the photometric reconstruction loss, a source image is warped into a target image frame using the predicted scene depths and the inter-frame pose change. The per-pixel reconstruction error is computed by comparing the target image to the reconstructed image; networks are trained to minimize the loss through gradient descent. 

A limitation of the photometric reconstruction loss is that it can only be used to train depth and egomotion networks that produce \textit{unscaled} predictions. Furthermore, the predictions are \textit{scale inconsistent}: different inputs produce depth and egomotion predictions with a varying scale factor, since there is nothing in the loss formulation that encourages independent predictions to have the same scale. To address scale inconsistency, recent works \cite{bian2019unsupervised,zhao2020masked} have proposed to enforce a global scale factor using a depth consistency loss. Despite producing scale-consistent estimates, these losses cannot resolve metric scale. 

To resolve scale in a self-supervised system, Godard et al.\ \cite{godard2017unsupervised} introduce a left-right consistency loss that uses stereo image pairs with a fixed (and known) baseline distance. However, despite being ``self-supervised'' in nature, stereo consistency losses cannot be used for retraining when only a single camera is available. To the best of the authors' knowledge, there is presently no self-supervised loss function that enforces metric scale for monocular systems. We formulate a loss function that is able to do so, by making use of the known camera height relative to the ground plane. Although the camera height may be considered as ground truth information, this quantity often remains available at test time, which facilitates online retraining.

The work most similar to our own is DNet \cite{xue2020toward}, which uses an online technique to estimate the scale factor of its learning-based depth and egomotion networks by detecting the ground plane. DNet requires the presence of a visible ground plane at test time to resolve the scale of the depth and egomotion estimates, while we embed information about metric scale during the training procedure and thus do not require a ground plane at test time. This simplifies scale recovery and makes our predictions less prone to failure at test time when the ground plane is not visible or is incorrectly detected.

\subsection{Application of Learning-Based VO}

Although we use a basic loss formulation and network structure to showcase our proposed scale recovery loss, we note that other learning-based methods \cite{yang2020d3vo,greenemetrically,czarnowski2020deepfactors,wagstaff2020self} produce state-of-the-art accuracy for monocular VO by incorporating learned predictions within classical (probabilistic and optimization-based) frameworks. Since these methods currently require stereo images or ground truth during training, they are not fully self-supervised in a monocular setting and cannot take advantage of online retraining; our proposed scale recovery method could easily be incorporated into these systems to maintain their ability to produce scaled depth and egomotion predictions while obviating the need for stereo images or ground truth poses.

\section{Approach}

In order to resolve metric scale in a self-supervised manner, we rely on three separate networks for depth estimation, egomotion estimation, and ground plane segmentation. Below, we introduce the depth and egomotion networks, along with the self-supervised loss formulation used to jointly train them. Then, the formulation of the plane segmentation loss (and network) and the scale recovery loss are discussed.

\subsection{Self-Supervised Depth and Egomotion Networks}

Our depth network is based on a U-Net \cite{Ronneberger:2015} encoder-decoder network, which takes as input a target image $\mathbf{I}_{t}$ and outputs a dense (per-pixel) depth prediction $\hat{\mathbf{D}}_{t}$. The egomotion network takes as input a target image $\mathbf{I}_t$ and a nearby source image $\mathbf{I}_s$, and outputs $\hat{\mathbf{T}}_{t,s}$, the estimated $\LieGroupSE{3}$ pose change between image frames. The primary loss term used for jointly training the depth and egomotion networks is a photometric reconstruction loss. The source image $\mathbf{I}_{s}$ can be differentiably warped to a target image $\mathbf{I}_{t}$ to produce the reconstructed image $\hat{\mathbf{I}}_{{t}}$ using a spatial transformer network \cite{Jaderberg:2015},
\begin{align}
\hat{\mathbf{I}}_{{t}} = ST(\mathbf{I}_{s}, \hat{\mathbf{D}}_{t}, \hat{\mathbf{T}}_{t,s}, f_u, f_v, c_u, c_v),
\end{align}
where the last four inputs are the known camera intrinsic parameters. The photometric reconstruction loss is the $L_1$ error of the reconstructed image (i.e., the `ground truth' is the target image):
\begin{align}
\mathscr{L}_{\text{L}_1} = \left| \hat{\mathbf{I}}_{{t}}(u,v) - \mathbf{I}_{{t}}(u,v) \right|.
\end{align}
A structural similarity loss \cite{wang2004image} is used in conjunction with the $L_1$ loss (balanced by $\alpha\in[0,1]$) to produce the overall photometric reconstruction loss:
\begin{equation}
\mathscr{L}_{\text{P}} = (1-\alpha)\mathscr{L}_{\text{L}_1} + \alpha\mathscr{L}_\text{SSIM}.
\end{equation}
An inverse depth smoothness term \cite{godard2017unsupervised} ensures that the gradients (in the $x$ and $y$ directions) of the inverse depth prediction agree with the image gradients:
\begin{equation}
\mathscr{L}_\text{S} = \sum_{i\in\{x,y\}}\AbsoluteValue{\partial_i\left(\frac{1}{ \hat{\mathbf{D}}_{t}(u,v)}\right)}e^{-\Norm{\partial_iI_t(u,v)}}.
\end{equation}
\noindent To improve scale consistency, we employ the loss from \cite{bian2019unsupervised}, which ensures that the source depths, when transformed to the target frame using the predicted pose change (becoming $\hat{\mathbf{D}}'_{s}$), are consistent with the target depths:
\begin{align}
\label{eq:depthconsist}
\mathscr{L}_{DC} = \frac{\AbsoluteValue{\hat{\mathbf{D}}'_{s}(u,v) - \hat{\mathbf{D}}_{t}(u,v)}}{\hat{\mathbf{D}}'_{s}(u,v) + \hat{\mathbf{D}}_{t}(u,v)}.
\end{align}

\noindent Finally, we implement a pose consistency loss to ensure that the `forward' and `inverse' inter-frame translation predictions are consistent with each other:
\begin{align}
\mathscr{L}_\text{PC} = \AbsoluteValue{ \hat{\mathbf{t}}_{t,s} - \hat{\mathbf{t}}_{s,t}}.
\end{align}
The complete image-wise loss function consists of all of the loss terms above:
\begin{align}
\label{eq:loss_terms}
\mathscr{L}_\text{base} &= \sum_{u,v}\left(\lambda_\text{P}\mathscr{L}_\text{P} + \lambda_\text{S}\mathscr{L}_\text{S} + \lambda_\text{DC}\mathscr{L}_\text{DC}\right) + \lambda_\text{PC}\mathscr{L}_\text{PC}.
\end{align}

\noindent Our baseline system is trained with this loss to produce \textit{unscaled} depth and egomotion estimates. Next, we discuss how we augment this system by incorporating scale recovery into the training procedure.

\begin{figure*}[]
	\centering
	\begin{subfigure}[]{0.245\textwidth}
		\includegraphics[width=\textwidth]{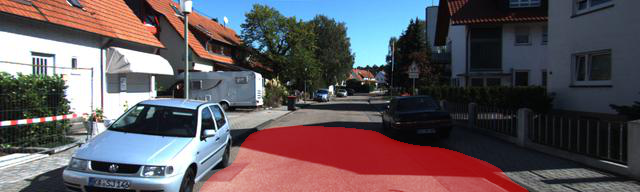}
		\label{fig:opt02}
	\end{subfigure}
	\begin{subfigure}[]{0.245\textwidth}
		\includegraphics[width=\textwidth]{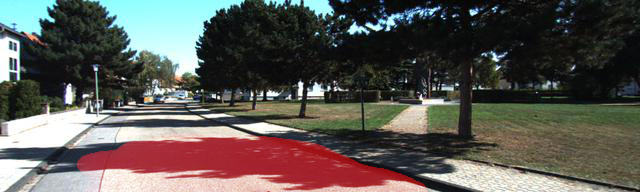}
		\label{fig:opt02}
	\end{subfigure}
	\begin{subfigure}[]{0.245\textwidth}
		\includegraphics[width=\textwidth]{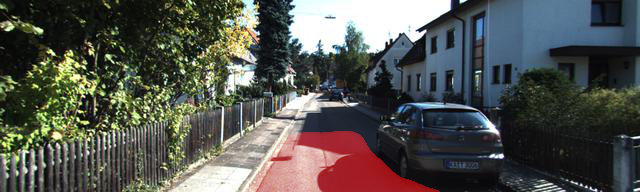}
		\label{fig:opt02}
	\end{subfigure}
	\begin{subfigure}[]{0.245\textwidth}
		\includegraphics[width=\textwidth]{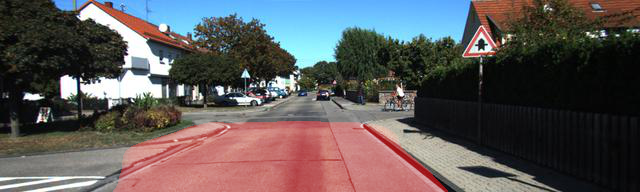}
		\label{fig:opt02}
	\end{subfigure} \\
	\vspace{-0.3cm}
	\begin{subfigure}[]{0.245\textwidth}
		\includegraphics[width=\textwidth]{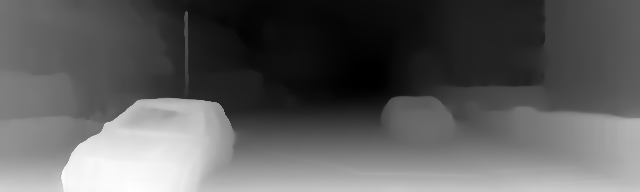}
		\label{fig:opt05}
	\end{subfigure} 
	\begin{subfigure}[]{0.245\textwidth}
		\includegraphics[width=\textwidth]{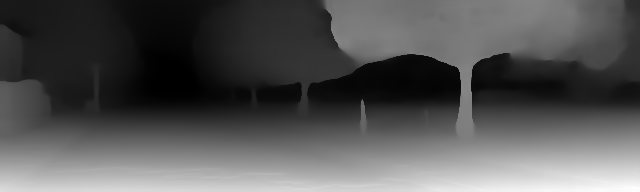}
		\label{fig:opt05}
	\end{subfigure} 
	\begin{subfigure}[]{0.245\textwidth}
		\includegraphics[width=\textwidth]{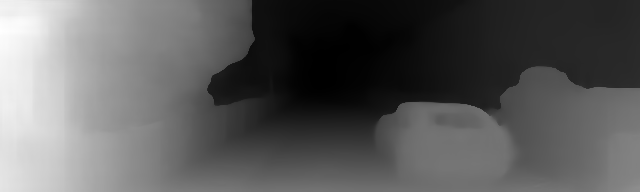}
		\label{fig:opt05}
	\end{subfigure} 
	\begin{subfigure}[]{0.245\textwidth}
		\includegraphics[width=\textwidth]{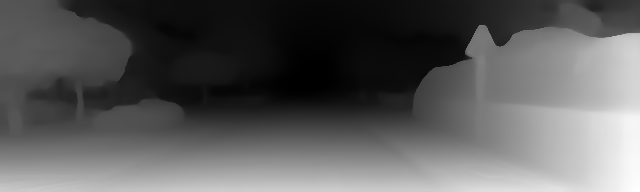}
		\label{fig:opt05}
	\end{subfigure} \\
	\vspace{-2mm}
	\caption{Examples of the plane segmentation masks (top row) and scene depth predictions (bottom row) produced by our plane segmentation and depth networks, respectively. The images are from KITTI sequence \texttt{05}.}
	\label{fig:planes_and_depths}
	\vspace{-0.3cm}
\end{figure*}

\subsection{Self-Supervised Ground Plane Segmentation}
\label{sec:approach_ground_plane_seg_network}

In our scale recovery approach, we compute the per-image scale factor of the depth predictions by observing the difference between the measured camera height and the known camera height. A scale factor of unity (corresponding to the heights being equal) is enforced during training by incorporating a novel scale recovery loss.

To estimate the scale factor, we first compute the camera height over the local ground plane, and compare it to the known camera height. This requires the ground plane itself to be extracted from the image. To extract the ground plane, we use our own plane segmentation network. Alternatively, the drivable road region could be detected using an existing supervised framework \cite{teichmann2018multinet}, but we choose to implement our own self-supervised technique in order to facilitate retraining alongside the depth and egomotion networks. Our plane segmentation network takes as input a target RGB image and outputs a corresponding plane segmentation mask $\hat{\mathbf{W}}_t$, whose per-pixel values $\hat{w}_t(u,v) \in [0,1]$ indicate the likelihood that each pixel is a ground plane `inlier'.

We train the ground plane segmentation network with a plane consistency loss. We assume that, for a given image, the lower, centre region contains the ground plane only.\footnote{Although this is a limiting assumption in general, it only applies to the training data, where we can be reasonably confident that the region consistently represents the road plane.} By computing the normal vector $\Normal_t$ and offset (i.e., the per-image camera height $h_t$) of the ground plane, we train our plane segmentation network to minimize a pixel-wise plane consistency loss $\mathscr{L}_\text{plane}$ over all pixels and images in the training dataset:
\begin{equation}
\label{eq:plane_consist}
\mathscr{L}_\text{plane} = \lambda_\text{plane}\hat{w}_t(u,v)\AbsoluteValue{h_t - \mathbf{p}_t(u,v)^T\Normal_t} -\lambda_\text{reg}\log \hat{w}_t(u,v).
\end{equation}
In order to minimize the first term, the per-pixel plane predictions $\hat{w}_t(u,v)$ must be small for pixels whose 3D coordinates $\mathbf{p}_t(u,v)$ do not lie on the ground plane. Since a trivial solution exists (outputting zero for all pixels in the image), the second term, a cross entropy regularization loss, enforces our segmentation mask outputs to be close to unity. The overall loss is minimized by training the network to accurately predict `inlier' plane pixels with high confidence while downweighting all other pixels. The two loss terms are balanced by the scalar weights $\lambda_\text{plane}$ and $\lambda_\text{reg}$.

In \cref{eq:plane_consist}, the per-image camera height $h_t$ is computed from the predefined ground plane region using a plane fitting procedure. For every ground plane pixel in the target image, the 3D coordinates $\mathbf{p}_t(u,v)$ are computed as 
\begin{align}
\label{eq:proj_3d}
\mathbf{p}_{t}(u,v) &=\hat{\mathbf{D}}_{t}(u,v)\begin{bmatrix} \frac{u - c_u}{f_u} & \frac{v - c_v}{f_v} & 1 \end{bmatrix}^T,
\end{align}
where a pre-trained depth network is used to estimate the scene depths $\hat{\mathbf{D}}_{t}$.\footnote{The pre-trained depth network that we use is unscaled---we use \cref{eq:loss_terms} to train this network using the same network structure and training procedure outlined in \Cref{sec:implementation_details}.} The 3D coordinates are stacked in $\mathbf{P}_t$, 
and the ground plane normal vector $\Normal_t$ is found by solving $\mathbf{P}_t^T\mathbf{n}_t = \mathbf{1}$ for $\mathbf{n}_t$. The unit normal to the plane is
\begin{equation}
\Normal_t = \frac{\mathbf{n}_t}{\Norm{\mathbf{n}_t}}.
\end{equation}
\noindent The estimated camera offset (i.e., the camera height relative to the plane) is then ${h}_t = \mathbf{P}_t^T \Normal_t$. We provide further details about the training procedure in \Cref{sec:implementation_details}. \Cref{fig:planes_and_depths} provides several examples of the plane segmentation network outputs. We note that our approach does assume local planarity of the ground, as other methods do. This assumption is only required for the training data, however, and images that break the assumption could  potentially be omitted.

\subsection{Scale Recovery Loss Formulation}

With knowledge of the ground plane (from our plane segmentation network), the most trivial way to enforce metric scale is to extract the estimated camera height $\hat{h}_t$, and enforce it to be similar to the known camera height, $h_{gt}$:
\begin{equation}
\label{eq:cam_height_loss}
\mathscr{L}_{cam} = |\hat{h}_{t} - h_{gt}|.
\end{equation}
The estimated camera height is determined through weighted least squares, where the ground plane normal vector is found by minimizing
\begin{equation}
L_t = \frac{1}{2}(\mathbf{P}_t^T\mathbf{n}_t - \mathbf{1})^T \mathbf{W}_t^{-1} (\mathbf{P}_t^T\mathbf{n}_t - \mathbf{1}), 
\end{equation}
\noindent where $\mathbf{P}_t \in\Real^{3\times HW} $ are the stacked 3D coordinates for every pixel in the image, $\mathbf{W}^{-1}_t \in\Real^{HW\times HW}$ is a diagonalized matrix of plane segmentation network outputs ($\hat{\mathbf{W}}_t$), and $\mathbf{1}$ is a vector of ones (of size $HW\times 1$). The least squares solution is:
\begin{equation}
\mathbf{n}_t = (\mathbf{P}_t\mathbf{W}_t^{-1}\mathbf{P}_t^T)^{-1}(\mathbf{P}_t\mathbf{W}_t^{-1}\mathbf{1}^T).
\end{equation}
\noindent The estimated camera height is a weighted average of the offset (relative to the plane) of all 3D coordinates:
\begin{equation}
\hat{h}_{t} = \frac{1}{\sum_{u,v}\hat{w}_t(u,v)}\sum_{u,v} \hat{w}_t(u,v)\,\mathbf{p}_t(u,v)^T \Normal_t.
\label{eq:camera_height}
\end{equation}
\noindent Importantly, since a weighted least squares approach is used, \cref{eq:cam_height_loss} becomes differentiable with respect to the depth predictions of \textit{every} pixel in the image (the current depth prediction is used to construct the 3D coordinates, $\mathbf{p}_t$, for each pixel). This link allows for our depth network weights to be updated through gradient descent by minimizing \cref{eq:cam_height_loss}, which metrically scales the depth predictions. We found, however, that there are issues that make \cref{eq:cam_height_loss} unsuitable in practice. Namely, because the loss is primarily a function of the ground plane pixels (since off-plane plane segmentation weights are generally close to zero), the ability of the depth network to properly resolve scale over the whole scene is limited. Instead of scaling all of the depth predictions, we found that the ground plane depth would erroneously `sink' as it became scaled, while other components of the scene remained unchanged.

To avoid the problem above, we propose an alternative loss that enforces metric depth (i.e., a scale factor of unity) by affecting all image pixels equally. Rather than directly comparing $\hat{h}_{t}$ to $h_{gt}$, as in \cref{eq:cam_height_loss}, we can compute an image-specific scale factor $s_t = \frac{h_{gt}}{\hat{h}_{t}}$, and generate per-pixel `depth scaling' targets
\begin{equation}
\label{eq:depth_labels}
\mathbf{D}^\text{DS}_t(u,v) = s_t \hat{\mathbf{D}}_t(u,v),
\end{equation}
\noindent which can be directly applied in a depth scaling loss:
\begin{equation}
\label{eq:depth_proxy}
\mathscr{L}_\text{DS} = \frac{|\hat{\mathbf{D}}_t(u,v) - \mathbf{D}^\text{DS}_t(u,v)|}{\mathbf{D}^\text{DS}_t(u,v)}.
\end{equation}
To enforce proper depth rescaling, all gradients associated with the target depth $\mathbf{D}^\text{DS}_t$ are removed (e.g., through $\mathbf{D}^\text{DS}_t$\texttt{.detach()} in \texttt{PyTorch}); this forces the network to update \textit{all} pixel depths, instead of only updating the ground plane pixels, because the only way to minimize \cref{eq:depth_proxy} is to update the network such that the estimated scale factor approaches unity.  The denominator in \cref{eq:depth_proxy} normalizes the per-pixel depth values to prevent large depths from dominating the loss function.  \Cref{fig:depth_scaling} illustrates how our depth scaling loss is applied.

We use the same technique (see \Cref{fig:trans_scaling}) to define a `translation scaling' loss,
\begin{equation}
\label{eq:trans_proxy}
\mathscr{L}_\text{TS} = \AbsoluteValue{\hat{\mathbf{t}}_{t, s} - \mathbf{t}^\text{TS}_{t,s}},
\end{equation}
\noindent where $\mathbf{t}^\text{TS}_{t,s} = s_t\hat{\mathbf{t}}_{t,s}$\texttt{.detach()}. We find that applying both scaling loss terms improves stability during training and causes the estimated scale factor to converge to unity more quickly. By combining our scale recovery loss with the baseline loss, our (per-sample) overall loss becomes
\begin{align}
\label{eq:proposed_loss}
\mathscr{L} = \mathscr{L}_\text{base} + \sum_{u,v}\left(\lambda_\text{DS}\mathscr{L}_\text{DS}\right) + \lambda_\text{TS}\mathscr{L}_\text{TS}.
\end{align}
\noindent By incorporating these scale recovery loss terms (balanced by $\lambda_\text{DS}$ and $\lambda_\text{TS}$), the scale factor will converge towards unity while the original loss terms are minimized.
\subsection{Implementation Details}
\label{sec:implementation_details}

We implemented our three networks in \texttt{PyTorch} \cite{paszke:2017}. The input to the depth and plane segmentation networks is a single RGB image, and the outputs are the depth and plane predictions, respectively. The depth and plane segmentation network encoder blocks were initialized with a pre-trained ResNet18 \cite{he2016deep} model, while the convolutional layers of the decoder blocks were set to the default initialization in \texttt{PyTorch}. The input to the egomotion network is two concatenated RGB images (a source and target image), as well as the optical flow between frames.\footnote{See the supplementary material (available in our open source repository) for additional network, training, and experimental details.} 

\begin{figure}[t]
	\centering
	\includegraphics[width=0.9\columnwidth]{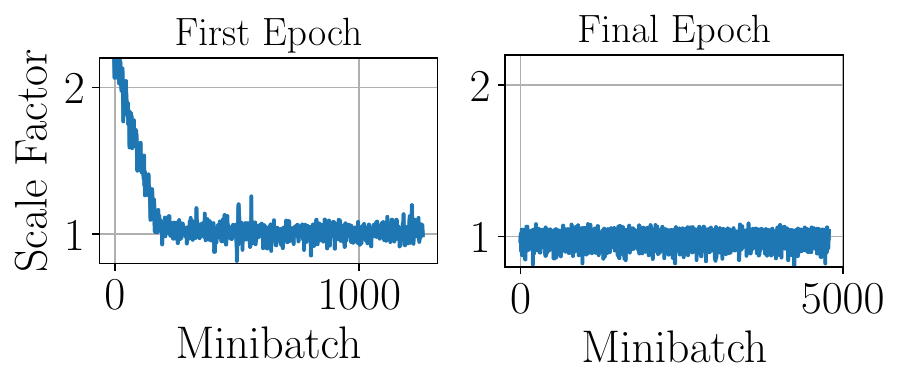}
	\vspace{-0.2cm}
	\caption{Scale factor convergence during training. Within one epoch of training on the KITTI dataset the scale factor effectively converges to unity.}
	\label{fig:scale_factor_convergence}
	\vspace{-0.4cm}
\end{figure}

We initially pretrained our models on the Oxford RobotCar  \cite{RobotCarDatasetIJRR} dataset. We subsequently trained the models for 25 epochs on the KITTI odometry dataset using the Adam optimizer \cite{Kingma:2014} with a minibatch size of six and a learning rate of $1\times10^{-4}$ that was reduced by half after every four epochs. The model that resulted in the lowest validation loss was selected for evaluation on a held-out test set. \Cref{fig:scale_factor_convergence} illustrates the convergence of the scale factor during training; the scale factor converges within 500 minibatch iterations. Over time, as illustrated by the final epoch in the training plot, the scale factor becomes more consistent, and is generally very close to unity.

\section{Experiments}
\label{sec:experiments}

We conducted several experiments to verify that our scale recovery method is able to produce depth and pose estimates that are metrically scaled with an accuracy comparable to existing scale recovery methods. Unlike these existing methods, however, our method is fully self-supervised and does not require stereo images or ground truth. We show that our scale recovery loss, by promoting scale consistency during training, is able to improve the overall VO accuracy compared with online scale recovery methods such as DNet \cite{xue2020toward}. Finally, we demonstrate how our loss formulation is well suited for online retraining to improve VO accuracy in new environments. For these experiments we used the KITTI Odometry \cite{Geiger:2013} and Oxford RobotCar \cite{RobotCarDatasetIJRR} datasets.

\subsection{Scale Factor Evaluation}
\label{sec:scale_factor_eval}

Our experimental results demonstrate that the scale recovery loss is able to accurately resolve the metric scale factor. We compared our method with two existing loss functions that are used to resolve scale: a pose supervision loss and a (stereo image) left-right consistency loss \cite{godard2017unsupervised}. To implement these two techniques, we directly replaced our scale recovery loss with the alternate loss function and trained the depth and egomotion networks from scratch. No changes were made to the training procedure\footnote{For the pose supervision method, we omit the odometry sequences (\texttt{11-21}) because no pose labels are available.} or the network structures, other than balancing the additional loss term with the existing loss terms by appropriately tuning its weighting factor.

\begin{figure}[]
	\centering
	\begin{subfigure}[]{0.49\columnwidth}
		\centering
		\includegraphics[width=\textwidth]{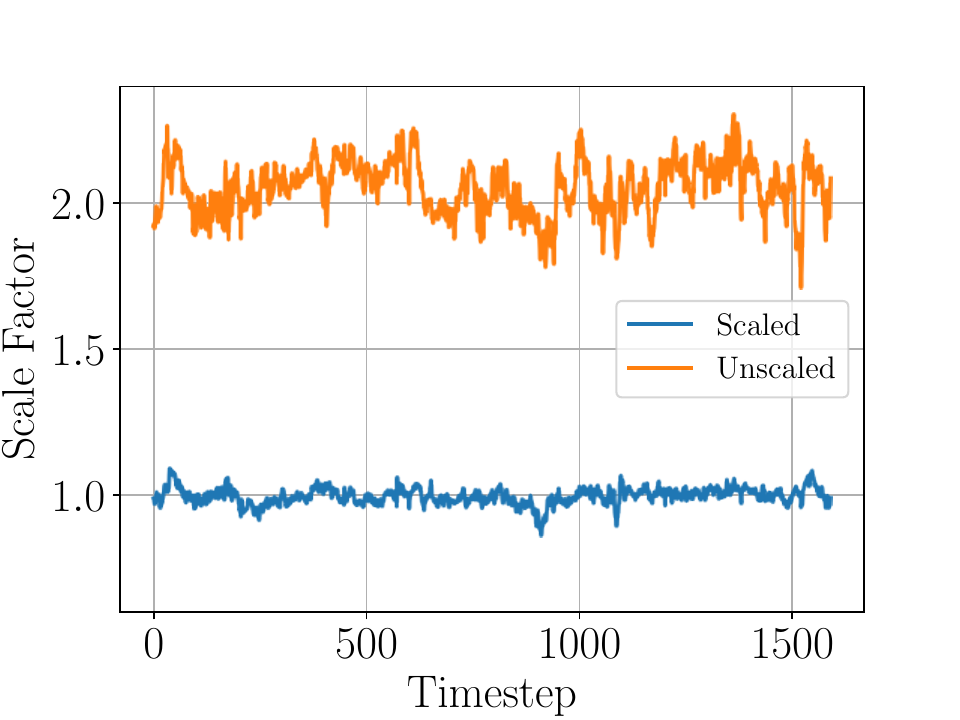}
		\vspace{-0.4cm}
		\caption{Seq. \texttt{09} scale factor.}
		\label{fig:scale05}
	\end{subfigure} 
	\begin{subfigure}[]{0.49\columnwidth}
		\centering
		\includegraphics[width=\textwidth]{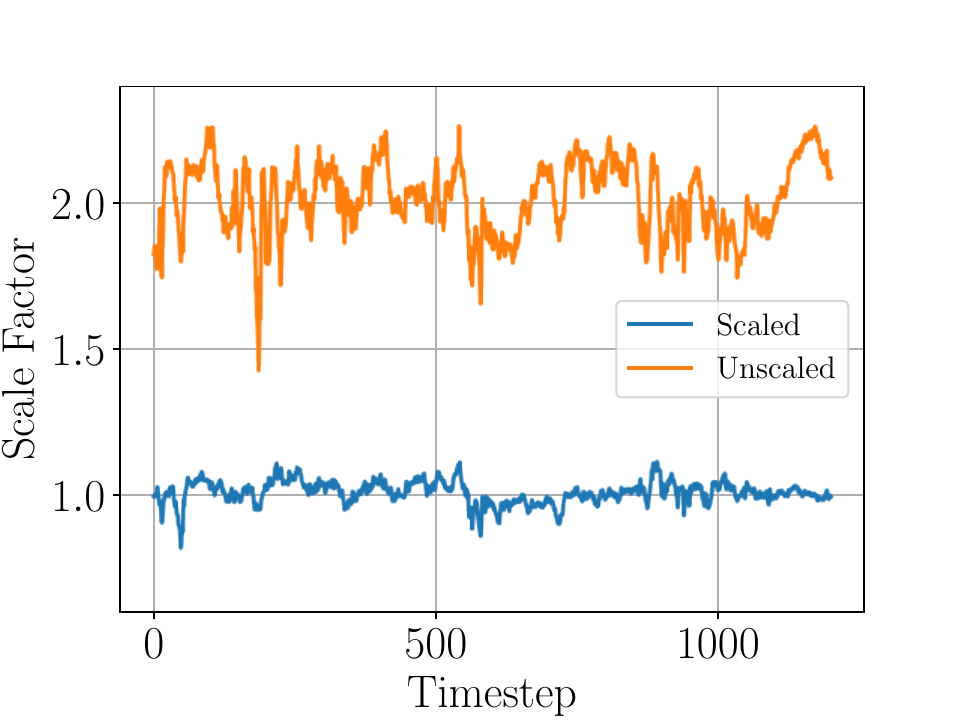}
		\vspace{-0.4cm}
		\caption{Seq. \texttt{10} scale factor.}
		\label{fig:scale10}
	\end{subfigure} 
	\caption{Visualization of the scale factor estimates for the KITTI test sequences, which indicate that our scale recovery loss promotes scale-consistent depth predictions with a scale factor close to unity.}
	\label{fig:mono-traj}
\end{figure}

To compare these three scale-resolving approaches, we estimated the scale factor of their depth predictions by extracting the ground plane using our plane segmentation network, and then computed the camera height with \cref{eq:camera_height}. We compared the estimated camera height $\hat{h}_{t}$ to the known camera height to determine the scale factor $s_t=\frac{h_{gt}}{\hat{h}_{t}}$ for every image frame within the test and validation sequences. We report the per-sequence mean scale factor in \cref{tab:scalefactor}. Comparing our scale recovery technique with the two alternate methods, the difference in scale factor is negligible. However, the alternative approaches require stereo images or ground truth information, while our algorithm requires knowledge of the camera height only (and only at training time). \Cref{fig:mono-traj} illustrates the estimated scale factor determined by our method and compares it with our baseline unscaled model. %
Lastly, we include an ablation study (\cref{tab:ablation}), which indicates that combining $\mathscr{L}_\text{DS}$ and $\mathscr{L}_\text{TS}$ results in a scale factor that is closest to unity, with the lowest standard deviation.
\begin{table}[]
	\centering
	\caption{Scale factor evaluation results. The average scale factors for three separate methods are reported for our KITTI test and validation sequences. A more accurate scale factor approaches unity.} 
	\label{tab:scalefactor}
	\begin{threeparttable}
		\begin{tabular}{cccc}
			\toprule
			Loss Type & \multicolumn{3}{c}{Scale Factor Mean (Std. Dev.)} \\ \toprule
			& \begin{tabular}[c]{@{}c@{}}Seq. 05\\ (val.)\end{tabular} & \multicolumn{1}{c}{\begin{tabular}[c]{@{}c@{}}Seq. 09\\ (test)\end{tabular}} & \multicolumn{1}{c}{\begin{tabular}[c]{@{}c@{}}Seq. 10\\ (test)\end{tabular}} \\ \midrule
			Pose Sup. & 0.976 (0.036) & 0.995 (0.042) & 1.012 (0.048)\\
			Stereo Consist. & 1.011 (0.032) & 1.012 (0.045) & 1.010 (0.045) \\
			Scale Recov. (Ours) & 1.004 (0.042) & 0.997 (0.028) & 1.009 (0.038) \\ \bottomrule
		\end{tabular}
	\end{threeparttable}
	\vspace{-0.3cm}
\end{table}
\begin{table}[]
	\vspace{-0.3cm}
	\centering
	\caption{Ablation study demonstrating the effect of our scale recovery loss terms.}
	\label{tab:ablation}
	\begin{threeparttable}
		\begin{tabular}{ccll}
			\toprule
			Loss Type & \multicolumn{3}{c}{Mean Scale Factor (Std. Dev.)} \\ \toprule
			& \begin{tabular}[c]{@{}c@{}}Seq. 05\\ (val.)\end{tabular} & \multicolumn{1}{c}{\begin{tabular}[c]{@{}c@{}}Seq. 09\\ (test)\end{tabular}} & \multicolumn{1}{c}{\begin{tabular}[c]{@{}c@{}}Seq. 10\\ (test)\end{tabular}} \\ \midrule
			Unscaled & $1.99 (0.196)$  & $2.05 (0.090)$ & $2.015 (0.122)$\\
			$\mathscr{L}_\text{TS}$ & 1.020 (0.065) & 1.046 (0.042) & 1.036 (0.054) \\
			$\mathscr{L}_\text{DS}$ & $1.023 (0.061)$ & $1.046 (0.043)$  & $1.041 (0.057)$ \\
			$\mathscr{L}_\text{TS} + \mathscr{L}_\text{DS}$ & \textbf{1.004} (\textbf{0.042}) & \textbf{0.997} (\textbf{0.028}) & \textbf{1.009} (\textbf{0.038}) \\ \bottomrule
		\end{tabular}
	\end{threeparttable}
\end{table}

\subsection{Depth Evaluation}
\label{sec:depth_experiments}

We evaluated the quality of our scaled depth predictions using the data splits of Eigen et al.\ \cite{eigen2015predicting}, which separates the KITTI Odometry dataset into standard training, validation, and test splits. For training, we used the preprocessing of Zhou et al.\ \cite{zhou:2017} to remove stationary images, and trained using the same network and procedure from \cref{sec:implementation_details}.

For evaluation of the depth predictions for the 697 test images, we report the standard metrics in \cref{tab:deptheval}, alongside the results from other scale-consistent approaches \cite{bian2019unsupervised,zhao2020masked}, and the approach most similar to our own, DNet \cite{xue2020toward}. We include three versions of our network: our (unscaled) baseline network whose predictions have been rescaled using ground truth depth (the per-image scale factor is $s_t=\frac{median(\mathbf{D}_{t,gt})}{median(\mathbf{D}_{t,pred})}$), the same network whose predictions have been rescaled using the known camera height (the per-image scale factor is $s_t=\frac{h_{gt}}{\hat{h}_{t}}$), and our proposed scaled depth network trained with our scale recovery loss. Comparing these methods, we see that the accuracy of our proposed method is competitive with existing approaches, despite not requiring any form of scaling at test time. Furthermore, we note that incorporating our scale recovery loss at training time results in more accurate depth predictions than our unscaled baseline predictions that have been rescaled using the known camera height, or even using ground truth. We posit that our scale recovery loss better enforces depth consistency during training and, as a result, our network produces higher quality depth predictions. \cref{tab:ablation} and \Cref{fig:mono-traj} support this claim: the variance of the scale factor is significantly smaller when our scale recovery loss is incorporated, which indicates that the depth predictions are generally consistent.
\begin{table*}[t!]
	\centering
	\caption{Monocular depth prediction results on the Eigen test split \cite{eigen2015predicting}. ``GT'' scaling is a per-image scale factor correction using the ground truth depth, while ``camera height'' scaling only uses knowledge of the camera height over the ground plane.}
	\label{tab:deptheval}
	\begin{threeparttable}
		\begin{tabular}{ccc@{\hspace{0.9\tabcolsep}}ccccccc}
			\toprule
			Method & Img. Resolution & Scaling Method & \multicolumn{4}{c}{Error $\downarrow$} & \multicolumn{3}{|c}{Accuracy $\uparrow$} \\ \midrule
			& &  & Abs Rel & Sq Rel & RMSE & RMSE log & $\delta < 1.25$ & $\delta < 1.25^2$ & $\delta < 1.25^3$ \\ \midrule
			Bian et al. \cite{bian2019unsupervised} & $256\times832$ & GT$^1$ & 0.137 & 1.089 & 5.439 & 0.217 & 0.830 & 0.942 & 0.975 \\
			Zhao et al. \cite{zhao2020masked} & $128\times416$ & GT & 0.148 & 1.091 & 5.536 & 0.209 & 0.802 & 0.934 & 0.976 \\
			DNet \cite{xue2020toward} & $192\times640$ & GT & \textbf{0.113} & \textbf{0.864} & \textbf{4.812} &\textbf{ 0.191} &\textbf{ 0.877} & \textbf{0.960} & \textbf{0.981} \\
			Ours | no $\mathscr{L}_\text{TS} + \mathscr{L}_\text{DS}$ & $192\times640$ & GT & 0.130 & 1.313 & 5.254 & 0.206 & 0.857 & 0.955 & 0.979 \\ \midrule
			DNet \cite{xue2020toward} & $192\times640$ & Cam. Height & 0.118 & 0.925 & 4.918 & 0.199 & 0.862 & 0.953 & 0.979 \\
			Ours | no $\mathscr{L}_\text{TS} + \mathscr{L}_\text{DS}$ & $192\times640$ & Cam. Height & 0.155 & 1.657 & 5.615 & 0.236 & 0.809 & 0.924 & 0.959 \\ \midrule
			Ours & $192\times640$ & None & 0.123 & 0.996 & 5.253 & 0.213 & 0.840 & 0.947 & 0.978 \\ \bottomrule
		\end{tabular}
		\begin{tablenotes}
			\item[1] A constant scale factor correction is applied to all frames in the sequence, as opposed to the per-frame ``GT'' scaling that other methods use for evaluation.
		\end{tablenotes}
	\end{threeparttable}
	\vspace{-0.4cm}
\end{table*}

\subsection{Visual Odometry Evaluation}
\label{sec:vo_experiments}

\begin{figure}[b!]
	\centering
	\vspace{-0.2cm}
	\includegraphics[width=0.85\columnwidth]{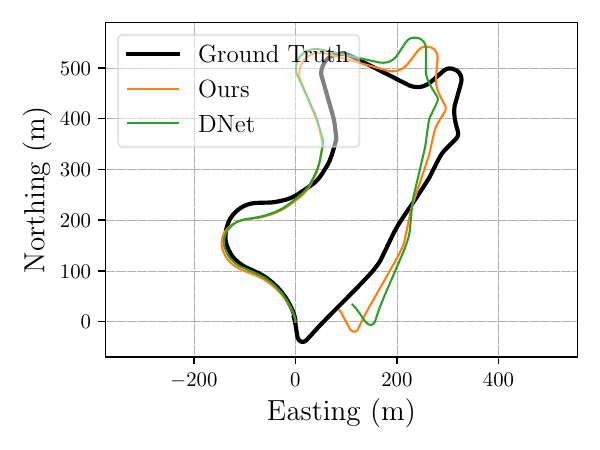}
	\vspace{-0.2cm}
	\caption{Top-down view of sequence \texttt{09}, comparing the accuracy of our method with the DNet rescaling method.}
	\label{fig:vo_seq_09}
\end{figure}

The aim of this experiment was to show that our proposed scale recovery method is able to produce metrically scaled egomotion estimates on the KITTI dataset. To generate the trajectory estimates for each sequence, the available $\LieGroupSE{3}$ inter-frame egomotion predictions were compounded together. Our evaluation criteria are the standard translation and rotation segment errors, averaged across all subsequences of length $\{100, 200, \dots, 800\}$ metres. Results are reported in \Cref{tab:vo_results}, and we include visualization of the trajectories in \Cref{fig:vo_seq_09} and in our supplementary material. 

We benchmarked the VO accuracy of our method by comparing it with several (monocular) alternatives. First, to illustrate the problem of scale drift, we benchmarked against \texttt{ORB-SLAM2} (without loop closure). Second, we evaluated performance against two existing learning-based VO methods \cite{bian2019unsupervised,zhao2020masked} that promote scale consistency (\hspace{-0.25em} \cite{bian2019unsupervised} uses the same depth consistency loss as us and \cite{zhao2020masked} uses a variant of this loss). Third, we compared our system against DNet \cite{xue2020toward}. Since the authors of \cite{xue2020toward} did not report any VO results, we adopted their open-source scale recovery technique and used it to rescale our baseline, unscaled network. We followed the same procedure as DNet to estimate the scale factor online: we computed the median camera height and determined the scale factor as $s_t = \frac{ h_{gt} } { \hat{h}_{t}} $. The per-image scale factor was then used to rescale the egomotion estimates from our baseline network.

\Cref{tab:vo_results} lists all mean translation and rotation segment errors for the aforementioned approaches. We see that \texttt{ORB-SLAM2} suffers from scale-inconsistency in seq. \texttt{09} compared to the other techniques. Our methods have similar accuracy to the other scale-consistent approaches \cite{bian2019unsupervised,zhao2020masked}, yet do not require any form of ground truth at test time to produce metrically-scaled predictions.

Interestingly, although DNet and our method use the same information to resolve metric scale, the incorporation of metric information during training (through our scale recovery loss) produces more accurate translation predictions compared with online rescaling. To confirm this, we removed the effect of orientation error by pairing the learned translation predictions with ground truth orientation, and additionally report these results in \Cref{tab:vo_results}. Similarly, our scaled translation predictions are more accurate than the unscaled predictions that are rescaled online. We hypothesize that this improvement comes from using our scale recovery loss during training, which improves the accuracy of our depth network by promoting scale-consistency (see \cref{sec:depth_experiments}). 

\begin{table}[b]
	\centering
	\caption{Visual odometry results on the KITTI dataset.}
	\label{tab:vo_results}
	\begin{threeparttable}
		\begin{tabular}{cc@{\hspace{0.9\tabcolsep}}c@{\hspace{0.9\tabcolsep}}c@{\hspace{0.9\tabcolsep}}c@{\hspace{0.9\tabcolsep}}c@{\hspace{0.9\tabcolsep}}c@{\hspace{0.9\tabcolsep}}}
			\toprule
			Method& \begin{tabular}[c]{@{}c@{}}Scaling\\ Method\end{tabular} & \multicolumn{2}{c}{Seq. \texttt{09}} & \multicolumn{2}{c}{Seq. \texttt{10}} \\ \midrule
			&  & \begin{tabular}[c]{@{}c@{}}trans. \\ (\%)\end{tabular} & \begin{tabular}[c]{@{}c@{}}rot.\\ ($^{\circ}$/100m)\end{tabular} & \begin{tabular}[c]{@{}c@{}}trans. \\ (\%)\end{tabular} & \begin{tabular}[c]{@{}c@{}}rot.\\ ($^{\circ}$/100m)\end{tabular} \\ \midrule
			\multicolumn{6}{c}{With Predicted Orientation} \\ \midrule
			ORB-SLAM & GT$^1$ & 15.30 & \textbf{0.26} & \textbf{3.68} & \textbf{0.48} \\
			Bian et al. \cite{bian2019unsupervised} & GT$^1$ & 8.24 & 2.19 & 10.7 & 4.58 \\
			Zhao et al. \cite{zhao2020masked} & GT$^1$ & 8.13 & 2.64 & 9.74 & 3.58 \\	
			Ours (no $\mathscr{L}_\text{TS} + \mathscr{L}_\text{DS}$)& DNet \cite{xue2020toward} & 7.23 & 1.91 & 13.98 & 4.07 \\
			Ours & None & \textbf{5.93} & 1.67 & 10.54 & 4.03 \\ \midrule
			\multicolumn{6}{c}{With Ground Truth Orientation} \\ \midrule
			Ours (no $\mathscr{L}_\text{TS} + \mathscr{L}_\text{DS}$) & DNet \cite{xue2020toward} & 5.14 & 0 & 9.67 & 0 \\
			Ours & None & \textbf{3.63} & 0 & \textbf{6.14} & 0 \\	
			\bottomrule
		\end{tabular}
		\begin{tablenotes}
			\item[1] A constant scale factor correction is applied to all frames in the sequence.
		\end{tablenotes}
	\end{threeparttable}
\end{table}
\subsection{Online Retraining Evaluation}

This experiment demonstrated how our self-supervised loss formulation is able to account for out-of-distribution data through online network retraining. Data collected within new environments can be used to update the model parameters, allowing the networks to adapt to changing surroundings. We examined domain adaptation by first training our system on the Oxford RobotCar dataset and then evaluating VO accuracy on the KITTI dataset. Note that we used the plane segmentation network trained on KITTI for retraining with our scale recovery loss. \Cref{tab:retraining_results} lists the domain adaptation results. We first trained our depth and egomotion networks on a subset of the Oxford RobotCar dataset using the same training procedure, network structure, and hyperparameters as for the KITTI experiments (see \Cref{sec:implementation_details}). As expected, due to the difference in the images between datasets (e.g., camera parameters and height, as well as significant changes in scene structure/illumination), our depth and egomotion networks trained on the RobotCar dataset did not perform well on the KITTI dataset. Following this, we retrained the networks for a single epoch with KITTI data (using the KITTI training sequences), without changing any parameters (with the exception of setting the camera height back to $1.70$ meters). To gauge the effectiveness of our scale recovery loss, retraining was carried out twice, once with the scale recovery loss and once without. By retraining the models for a single epoch on KITTI using the \cref{eq:loss_terms} baseline (unscaled) loss, the overall accuracy improved, but the scale factor did not converge to unity. In contrast, when retraining with \cref{eq:proposed_loss}, the network adapted to the KITTI environment \textit{and} produced metric predictions. 

\begin{table}[t!]
	\centering
	\caption{Results from our retraining experiment. A network pre-trained on the Oxford RobotCar dataset is evaluated on the KITTI dataset. By retraining on KITTI with our scale recovery loss, VO accuracy is significantly improved.}
	\label{tab:retraining_results}
	\begin{tabular}{cc@{\hspace{0.9\tabcolsep}}c@{\hspace{0.9\tabcolsep}}c@{\hspace{0.9\tabcolsep}}c@{\hspace{0.9\tabcolsep}}}
		\toprule
		Method & \multicolumn{2}{c}{Seq. \texttt{09}} & \multicolumn{2}{c}{Seq. \texttt{10}}  \\ \midrule
		& \begin{tabular}[c]{@{}c@{}}trans.\\ (\%)\end{tabular} & \begin{tabular}[c]{@{}c@{}}deg.\\ ($^{\circ}$/100m)\end{tabular} & \begin{tabular}[c]{@{}c@{}}trans.\\ (\%)\end{tabular} & \begin{tabular}[c]{@{}c@{}}deg.\\ ($^{\circ}$/100m)\end{tabular} \\ \midrule
		Original & 34.51 & 9.23 & 36.38 & 12.01 \\
		Retrained (no $\mathscr{L}_\text{TS} + \mathscr{L}_\text{DS}$) & 34.99 & 2.25 & 32.53 & \textbf{3.45} \\
		\begin{tabular}[c]{@{}c@{}} Retrained (no $\mathscr{L}_\text{TS} + \mathscr{L}_\text{DS}$) \\ with DNet Rescaling \end{tabular} & 7.73 & 2.25 & 12.48 & \textbf{3.45} \\
		Retrained (with $\mathscr{L}_\text{TS} + \mathscr{L}_\text{DS}$) & \textbf{6.93} & \textbf{1.70} & \textbf{8.82} & 3.54 \\ \bottomrule
	\end{tabular}
	\vspace{-0.3cm}
\end{table}
\section{Conclusion}
In this paper, we described how a novel scale recovery loss can be added to a self-supervised depth and egomotion estimation pipeline to produce metrically-scaled predictions. In contrast to alternative approaches (e.g., pose supervision or stereo consistency losses), our method only requires a stream of monocular images and a known camera height at training time. Notably, our networks can be retrained online, which significantly improves egomotion predictions for out-of-distribution images. Additionally, our loss enforces depth consistency during training, boosting overall egomotion estimation accuracy compared to a similar method that only recovers scale at test time. As future work, we plan to incorporate uncertainty into the scale recovery loss to account for camera height changes due to vertical motion (e.g., tilt) of the vehicle.
\vspace{-1mm}
\section*{Acknowledgments}
We gratefully acknowledge the contribution of NVIDIA Corporation, who provided the Titan X GPU used for this research through their Hardware Grant Program.

\vspace{-1mm}
\bibliographystyle{IEEEcaps}
\bibliography{example.bib}

\begin{thebibliography}{10}
\def\url#1{}
\csname url@rmstyle\endcsname
\providecommand{\newblock}{\relax}
\providecommand{\bibinfo}[2]{#2}
\providecommand\BIBentrySTDinterwordspacing{\spaceskip=0pt\relax}
\providecommand\BIBentryALTinterwordstretchfactor{4}
\providecommand\BIBentryALTinterwordspacing{\spaceskip=\fontdimen2\font plus
\BIBentryALTinterwordstretchfactor\fontdimen3\font minus
  \fontdimen4\font\relax}
\providecommand\BIBforeignlanguage[2]{{%
\expandafter\ifx\csname l@#1\endcsname\relax
\typeout{** WARNING: IEEEtran.bst: No hyphenation pattern has been}%
\typeout{** loaded for the language `#1'. Using the pattern for}%
\typeout{** the default language instead.}%
\else
\language=\csname l@#1\endcsname
\fi
#2}}

\bibitem{Aqel:2016}
M.~Aqel, M.~Marhaban, M.~Saripan, and N.~Ismail, ``Review of visual odometry:
  types, approaches, challenges, and applications,'' \emph{SpringerPlus},
  vol.~5, no.~1, p. 1897, 2016.

\bibitem{frost2016object}
D.~Frost, O.~K{\"a}hler, and D.~Murray, ``Object-aware bundle adjustment for
  correcting monocular scale drift,'' in \emph{Proc. IEEE Int. Conf. Robot.
  Autom. (ICRA)}, 2016, pp. 4770--4776.

\bibitem{frost2018recovering}
D.~Frost, V.~Prisacariu, and D.~Murray, ``Recovering stable scale in monocular
  SLAM using object-supplemented bundle adjustment,'' \emph{IEEE Trans. on
  Robots.}, vol.~34, no.~3, pp. 736--747, 2018.

\bibitem{yin2017scale}
X.~Yin, X.~Wang, X.~Du, and Q.~Chen, ``Scale recovery for monocular visual
  odometry using depth estimated with deep convolutional neural fields,'' in
  \emph{Proc. IEEE Int. Conf. Comput. Vis. (ICCV)}, 2017, pp. 5870--5878.

\bibitem{kreuzig2019distancenet}
R.~Kreuzig, M.~Ochs, and R.~Mester, ``DistanceNet: Estimating traveled distance
  from monocular images using a recurrent convolutional neural network,'' in
  \emph{Proc. IEEE/CVF Conf. Comput. Vision Pattern Recognition (CVPR)
  Workshop}, 2019.

\bibitem{Wang:2017}
S.~Wang, R.~Clark, H.~Wen, and N.~Trigoni, ``{DeepVO}: Towards end-to-end
  visual odometry with deep recurrent convolutional neural networks,'' in
  \emph{Proc. IEEE Int. Conf. Robot. Autom. (ICRA)}, 2017, pp. 2043--2050.

\bibitem{krueger2020out}
D.~Krueger, \emph{et~al.}, ``Out-of-distribution generalization via risk
  extrapolation (REx),'' in \emph{Proc. Int. Conf. Machine Learn. (ICML)},
  2021, pp. 5815--5826.

\bibitem{zhou:2017}
T.~Zhou, M.~Brown, N.~Snavely, and D.~Lowe, ``Unsupervised learning of depth
  and ego-motion from video,'' in \emph{Proc. IEEE/CVF Conf. Comput. Vision
  Pattern Recognition (CVPR)}, 2017, pp. 1851--1858.

\bibitem{xue2020toward}
F.~Xue, G.~Zhuo, Z.~Huang, W.~Fu, Z.~Wu, and M.~A. Jr, ``Toward hierarchical
  self-supervised monocular absolute depth estimation for autonomous driving
  applications,'' in \emph{Proc. Conf. IEEE/RSJ Int. Conf. Intell. Robot. Syst.
  (IROS)}, 2020, pp. 2330--2337.

\bibitem{mur2017orb}
R.~Mur-Artal and J.~Tard{\'o}s, ``ORB-SLAM2: An open-source SLAM system for
  monocular, stereo, and RGB-D cameras,'' \emph{IEEE Trans. Robot.}, vol.~33,
  no.~5, pp. 1255--1262, 2017.

\bibitem{song2015high}
S.~Song, M.~Chandraker, and C.~Guest, ``High accuracy monocular SfM and scale
  correction for autonomous driving,'' \emph{IEEE Trans. Pattern Anal. Mach.
  Intell.}, vol.~38, no.~4, pp. 730--743, 2015.

\bibitem{fanani2017multimodal}
N.~Fanani, A.~St{\"u}rck, M.~Barnada, and R.~Mester, ``Multimodal scale
  estimation for monocular visual odometry,'' in \emph{Proc. IEEE Intell. Veh.
  Symp. (IV)}, 2017, pp. 1714--1721.

\bibitem{zhou2019ground}
D.~Zhou, Y.~Dai, and H.~Li, ``Ground-plane-based absolute scale estimation for
  monocular visual odometry,'' \emph{IEEE Trans. Intell. Transp. Syst.},
  vol.~21, no.~2, pp. 791--802, 2019.

\bibitem{wang2018monocular}
X.~Wang, H.~Zhang, X.~Yin, M.~Du, and Q.~Chen, ``Monocular visual odometry
  scale recovery using geometrical constraint,'' in \emph{Proc. IEEE Int. Conf.
  Robot. Autom. (ICRA)}, 2018, pp. 988--995.

\bibitem{kitt2011monocular}
B.~Kitt, J.~Rehder, A.~Chambers, M.~Schonbein, H.~Lategahn, and S.~Singh,
  ``Monocular visual odometry using a planar road model to solve scale
  ambiguity,'' in \emph{Proc. Eur. Conf. Mobile Robot.}, 2011.

\bibitem{lee2015online}
B.~Lee, K.~Daniilidis, and D.~Lee, ``Online self-supervised monocular visual
  odometry for ground vehicles,'' in \emph{Proc. IEEE Int. Conf. Robot. Autom.
  (ICRA)}, 2015, pp. 5232--5238.

\bibitem{wang2019improving}
X.~Wang, D.~Maturana, S.~Yang, W.~Wang, Q.~Chen, and S.~Scherer, ``Improving
  learning-based ego-motion estimation with homomorphism-based losses and drift
  correction,'' in \emph{Proc. Conf. IEEE/RSJ Int. Conf. Intell. Robot. Syst.
  (IROS)}, 2019, pp. 970--976.

\bibitem{costante2018ls}
G.~Costante and T.~Ciarfuglia, ``LS-VO: Learning dense optical subspace for
  robust visual odometry estimation,'' \emph{{IEEE} Robot. Autom. Lett.},
  vol.~3, no.~3, pp. 1735--1742, 2018.

\bibitem{godard2019digging}
C.~Godard, O.~Aodha, M.~Firman, and G.~Brostow, ``Digging into self-supervised
  monocular depth estimation,'' in \emph{Proc. IEEE/CVF Conf. Comput. Vision
  Pattern Recognition (CVPR)}, 2019, pp. 3828--3838.

\bibitem{kumar2019fisheyedistancenet}
V.~Kumar, \emph{et~al.}, ``FisheyeDistanceNet: Self-supervised scale-aware
  distance estimation using monocular fisheye camera for autonomous driving,''
  in \emph{Proc. IEEE Int. Conf. Robot. Autom. (ICRA)}, 2020, pp. 574--581.

\bibitem{yang2020d3vo}
N.~Yang, L.~Stumberg, R.~Wang, and D.~Cremers, ``D3VO: Deep depth, deep pose
  and deep uncertainty for monocular visual odometry,'' in \emph{Proc. IEEE/CVF
  Conf. Comput. Vision Pattern Recognition (CVPR)}, 2020, pp. 1281--1292.

\bibitem{bian2019unsupervised}
J.~Bian, \emph{et~al.}, ``Unsupervised scale-consistent depth and ego-motion
  learning from monocular video,'' in \emph{Proc. Conf. Neural Inf. Process.
  Syst. (NeurIPS)}, 2019, pp. 35--45.

\bibitem{zhao2020masked}
C.~Zhao, G.~Yen, Q.~Sun, C.~Zhang, and Y.~Tang, ``Masked GANs for unsupervised
  depth and pose prediction with scale consistency,'' \emph{IEEE Trans. Neural
  Netw. Learn. Syst.}, 2020.

\bibitem{godard2017unsupervised}
C.~Godard, O.~Aodha, and G.~Brostow, ``Unsupervised monocular depth estimation
  with left-right consistency,'' in \emph{Proc. IEEE/CVF Conf. Comput. Vision
  Pattern Recognition (CVPR)}, 2017, pp. 270--279.

\bibitem{greenemetrically}
W.~Greene and N.~Roy, ``Metrically-scaled monocular SLAM using learned scale
  factors,'' in \emph{Proc. IEEE Int. Conf. Robot. Autom. (ICRA)}, 2020, pp.
  43--50.

\bibitem{czarnowski2020deepfactors}
J.~Czarnowski, T.~Laidlow, R.~Clark, and A.~Davison, ``Deepfactors: Real-time
  probabilistic dense monocular SLAM,'' \emph{{IEEE} Robot. Autom. Lett.},
  vol.~5, no.~2, pp. 721--728, 2020.

\bibitem{wagstaff2020self}
B.~Wagstaff, V.~Peretroukhin, and J.~Kelly, ``Self-supervised deep pose
  corrections for robust visual odometry,'' in \emph{Proc. IEEE Int. Conf.
  Robot. Autom. (ICRA)}, 2020, pp. 2331--2337.

\bibitem{Ronneberger:2015}
{Ronneberger, O. and Fischer, P. and Brox, T.}, ``{U-Net}: Convolutional
  networks for biomedical image segmentation,'' in \emph{Proc. Int. Conf. Med.
  Image Comput. Comput. Assist. Interv.}\hskip 1em plus 0.5em minus 0.4em\relax
  Springer, 2015, pp. 234--241.

\bibitem{Jaderberg:2015}
M.~Jaderberg, K.~Simonyan, A.~Zisserman, and K.~Kavukcuoglu, ``Spatial
  transformer networks,'' in \emph{Proc. Conf. Neural Inf. Process. Syst.
  (NeurIPS)}, 2015, pp. 2017--2025.

\bibitem{wang2004image}
Z.~Wang, A.~Bovik, H.~Sheikh, and E.~Simoncelli, ``Image quality assessment:
  from error visibility to structural similarity,'' \emph{IEEE Trans. Image
  Process.}, vol.~13, no.~4, pp. 600--612, 2004.

\bibitem{teichmann2018multinet}
M.~Teichmann, M.~Weber, M.~Zoellner, R.~Cipolla, and R.~Urtasun, ``MultiNet:
  Real-time joint semantic reasoning for autonomous driving,'' in \emph{Proc.
  IEEE Intell. Veh. Symp. (IV)}, 2018, pp. 1013--1020.

\bibitem{paszke:2017}
A.~Paszke, \emph{et~al.}, ``Automatic differentiation in {PyTorch},'' in
  \emph{Workshop on Automatic Differentiation, Conf. Neural Inf. Process. Syst.
  (NeurIPS)}, 2017.

\bibitem{he2016deep}
K.~He, X.~Zhang, S.~Ren, and J.~Sun, ``Deep residual learning for image
  recognition,'' in \emph{Proc. IEEE/CVF Conf. Comput. Vision Pattern
  Recognition (CVPR)}, 2016, pp. 770--778.

\bibitem{RobotCarDatasetIJRR}
W.~Maddern, G.~Pascoe, C.~Linegar, and P.~Newman, ``1 year, 1000 km: the Oxford
  RobotCar dataset,'' \emph{Int. J. Robot. Res. (IJRR)}, vol.~36, no.~1, pp.
  3--15, 2017.

\bibitem{Kingma:2014}
D.~Kingma and J.~Ba, ``Adam: A method for stochastic optimization,''
  \emph{arXiv preprint arXiv:1412.6980}, 2014.

\bibitem{Geiger:2013}
A.~Geiger, P.~Lenz, C.~Stiller, and R.~Urtasun, ``Vision meets robotics: the
  {KITTI} dataset,'' \emph{Int. J. Robot. Res. (IJRR)}, vol.~32, no.~11, pp.
  1231--1237, 2013.

\bibitem{eigen2015predicting}
D.~Eigen and R.~Fergus, ``Predicting depth, surface normals and semantic labels
  with a common multi-scale convolutional architecture,'' in \emph{Proc.
  IEEE/CVF Int. Conf. Comput. Vision (ICCV)}, 2015, pp. 2650--2658.

\end{thebibliography}
\end{document}